\documentclass{article}
\usepackage{arxiv}

\usepackage[utf8]{inputenc}
\usepackage[T1]{fontenc}
\usepackage{times}
\usepackage{geometry}
\usepackage{amsmath}
\usepackage{amssymb}
\usepackage{amsthm}
\usepackage{mathtools}
\usepackage{bm}

\usepackage{booktabs}
\usepackage{color, colortbl}
\usepackage{threeparttable}
\usepackage{nicefrac}
\usepackage{xcolor}
\usepackage{subcaption}
\usepackage{indentfirst}

\usepackage{algorithm}
\usepackage{algorithmic}

\usepackage[authoryear,round]{natbib}
\usepackage{hyperref}
\usepackage{url}

\DeclareMathOperator*{\argmax}{arg\,max}

\newtheorem{theorem}{Theorem}
\newtheorem{lemma}{Lemma}
\newtheorem{assumption}{Assumption}

\title{Coordinate Basis Linear Bandits with Sublinear Regret over Low-Rank Clusters}

\author{
Hantao Yang \\
  University of Science and Technology of China\\
  \texttt{yanghantao@mail.ustc.edu.cn} \\
   \And
Hong Xie\\
  University of Science and Technology of China\\
  \texttt{xiehong2018@foxmail.com} \\
  \And
Defu Lian\\
  University of Science and Technology of China\\
  \texttt{liandefu@ustc.edu.cn} \\
}

\begin{document}

\maketitle

\begin{abstract}
We study the problem of computational efficiency for linear bandits in high-dimensional settings with a finite arm set. In linear bandits, the increase in the dimension $d$ of the feature vectors leads to growing computational costs of $O(d^2)$ at each round of update. Traditional low-rank approximation methods such as SOFUL reduce computation via fixed-size matrix sketching, yet run the risk of incurring vacuous linear regret when the spectral tail of the data is heavy and the approximation rank is inadequately selected. To guarantee regret convergence and effectively reduce computational costs, we introduce a clustering mechanism and propose the Coordinate Basis Linear Bandit (CBL) algorithm. Our method preserves the full covariance information in each cluster by maintaining an exact orthonormal basis of the cluster's column space together with a low-dimensional coordinate representation, which guarantees robust sublinear regret without spectral-tail vulnerabilities. It performs cluster switching by assigning a sentinel for each cluster, and reduces per-round update computation to $O(\bar r^2)$ in the cluster's effective dimension $\bar r=\min(r,l)$, where $r$ is the intrinsic rank of the data and $l<d$ is a tunable rank bound.
\end{abstract}

\section{Introduction}
The stochastic linear contextual bandit is a fundamental framework for sequential decision-making under uncertainty, where an agent repeatedly selects an action from a finite set based on its associated context vector, and observes a noisy reward that is linear in the unknown parameter. This model has found widespread applications in personalized recommendation \citep{LinUCB}, online advertising \citep{10.1145/2505515.2514700}, and mobile health \citep{Tewari2017FromAT}, among others. Classical algorithms such as LinUCB \citep{ContextualBanditsLinearPayoff,OFUL} and linear Thompson Sampling \citep{Thompsonsamplingbandits} achieve near-optimal regret guarantees, but incur a prohibitive $O(d^2)$ per-round time complexity due to the need to maintain and invert a $O(d\times d)$ covariance matrix, where $d$ is the feature dimension.

As data scales have grown rapidly in recent years, both the feature dimension $d$ and the number of arms $N$ can be extremely large, rendering exact methods computationally infeasible. To reduce this update computation, recent works have turned to matrix sketching and low-rank approximation techniques, which maintain a low-dimensional approximation of the covariance matrix. SOFUL \citep{SOFUL} applies Frequent Directions \citep{FD} to compress the covariance matrix from dimension $d$ to $l < d$, reducing per-round complexity to $O(ld)$. The CBSCFD \citep{CBSCFD} algorithm improves upon SOFUL by leveraging the accumulated information of its truncated singular values, thereby achieving better regret performance. While these methods offer substantial computational speedups, they suffer from a critical drawback: the global sketching of the entire action-history covariance matrix inevitably introduces approximation errors that accumulate over time. This spectral distortion can inflate the confidence ellipsoid, degrade arm-selection accuracy, and in some cases lead to linear regret or even non-convergence, particularly when the covariance spectrum has a heavy tail. In other words, existing sketching methods sacrifice linear regression accuracy in favor of computational efficiency. However, this trade-off struggles to guarantee the convergence of regret, which may render them ineffective in systems that require stability guarantees. \citet{DBSLinUCB} improved upon SOFUL and proposed DBSLinUCB, which ensures sublinear regret through a multi-scale, adaptively adjusted sketch size. Although DBSLinUCB resolves the non-convergence issue of SOFUL, its theoretical guarantees rely on adaptive sketch-size growth and cannot be extended to the fixed-rank setting. Moreover, its computational efficiency is predicated on low-rank data; on full-rank data, it reverts to $O(d^2)$ complexity, losing the speed advantage of low-rank methods.

To achieve regret convergence in computation-constrained linear bandits, we propose the Coordinate Basis Linear Bandit (CBL). Instead of maintaining a lossy global sketch, CBL partitions the action set into clusters and, for each cluster, stores an \emph{exact} orthonormal basis $U_k$ of the cluster's column space together with the low-dimensional coordinates $Z_k$ of its arms. Because the basis spans the full range of the cluster's arm matrix and no direction is truncated, the regularized least-squares estimate and the confidence width computed in the coordinate space are pointwise identical to those of the exact $d$-dimensional LinUCB restricted to that cluster. Consequently, CBL retains the statistical accuracy of exact linear regression while reducing the per-round cost to $O(\bar r^2)$ in the cluster's effective dimension $\bar r=\min(r,l)$, where $r$ is the intrinsic rank of the data and $l<d$ is a tunable rank bound. To our knowledge, it is the first low-rank-based bandit strategy that reduces computational cost while guaranteeing regret convergence under a fixed rank bound. Our main contributions are as follows:
\begin{itemize}
\item \textbf{Coordinate-Basis Framework.} We propose an exact coordinate-basis framework for finite-action linear bandits. Unlike prior global sketching approaches, our method maintains a dedicated orthonormal basis and coordinate representation for each subset of arms, preserving the information integrity of each subset's column space, which serves as the foundation for our regret convergence guarantee.

\item \textbf{Regret Convergence and Computational Efficiency.} We prove that CBL achieves a regret bound of $\widetilde{O}(l\sqrt{N_{\mathcal{A}}^{l}T})$ and a per-round computational cost of $O(\bar r^2)$ with $\bar r=\min(r,l)$, where $r$ is the intrinsic rank of the data and $l$ is a tunable rank bound, excluding the additional overhead incurred during the warm-up phase, offering a clear efficiency gain when $l^2 < d$. Notably, unlike previous sketching-based methods, our regret bound is free of the spectral error term $\Delta_T$---the primary source of linear regret in existing approaches---thus ensuring rigorous regret convergence. Moreover, our theoretical framework remains valid under both a fixed rank bound and full-rank data settings.
\end{itemize}

\section{Related Work}
\textbf{Linear Bandits}
Contextual bandits generalize the finite-arm setting by allowing the exploitation of side information through linear models, thereby giving rise to the linear bandit framework. The study of regret minimization in linear bandits was first studied by \citet{645528.657779}, and has attracted extensive interest in the development of various algorithms \citep{LinearlyParameterized,ContextualBanditsLinearPayoff}. \citet{944919.944941} provided an early algorithm (LinRel) and theoretical analysis for this problem by introducing confidence-bound-based techniques, improving the regret bound from the previous $\widetilde{O}(T^{3/4})$ to $\widetilde{O}(\sqrt{T})$. Subsequently, \citet{Dani2008StochasticLO} extended the theoretical framework to more general compact action sets. \citet{LinUCB} proposed the LinUCB algorithm and successfully applied it to personalized recommendation. \citet{ContextualBanditsLinearPayoff} conducted further theoretical investigations of linear bandits by maintaining a regularized least-squares estimate of the unknown parameter and constructing a confidence ellipsoid centered on this estimate. By selecting the action that maximizes the upper confidence bound, their work demonstrated that LinUCB achieves a near-optimal regret of $\widetilde{O}(d\sqrt{T})$ under relatively mild assumptions. \citet{OFUL} further refined the analysis by developing a tighter confidence set based on self-normalized martingale inequalities, and introduced the classic OFUL strategy. \citet{Thompsonsamplingbandits} adapted Thompson Sampling \citep{Thompson1933ONTL} to linear bandits, and this method was later further refined by \citet{Abeille2016LinearTS}. Linear Thompson sampling randomly draws a parameter from the posterior distribution and selects the action that maximizes the expected reward under this sample. Its performance is comparable to that of OFUL up to logarithmic factors.

\textbf{Sketching and Low-Rank Bandits}
Sketching is a classic dimensionality reduction technique \citep{10.1145/3055399.3055431,pmlr-v80-andoni18a,BeyondJohnson} that has been introduced into linear bandits to reduce the computational cost of matrix inversion \citep{Yu_Lyu_King_2017,SOFUL}. \citet{FD} proposed Frequent Directions (FD), a sketching method with strong theoretical guarantees in streaming settings, making it particularly suitable for online environments. \citet{SOFUL} first combined FD with the classic OFUL algorithm in linear contextual bandits, yielding the SOFUL algorithm and reducing per-round update costs. Building on SOFUL, \citet{CBSCFD} developed a variant of FD for linear bandits that achieves a tighter regret bound. To overcome the potential non-convergence issue of traditional sketching methods, \citet{DBSLinUCB} proposed DBSLinUCB, a multi-scale sketching framework that adaptively adjusts the sketch size during learning, thus avoiding the linear regret trap of fixed-size sketches. Unlike previous single-scale approaches, DBSLinUCB dynamically distributes the sketch budget across blocks to guarantee sublinear regret without prior knowledge of the data spectrum. While these methods approximate the covariance matrix through sketches, our approach instead maintains an \emph{exact} low-dimensional representation: a per-cluster orthonormal basis of the column space, which preserves all covariance information within the cluster and therefore sidesteps the spectral-tail error entirely. While DBSLinUCB features an adaptive sketch size, our work focuses on the fixed-rank setting.

\section{Problem Setting}
\subsection{Notation and Definition}
In our paper, every d-dimensional vector is represented as a column vector. For matrix $\bm{A}$, $[\bm{A}]_k$ denotes the matrix formed by taking the first $k$ rows of matrix $\bm{A}$. For matrix $\bm{A}\in \mathbb{R}^{d_1\times d_2}$, the computed SVD of $\bm{A}$ is defined as $\bm{A} = \bm{U}\bm{\Sigma}\bm{V}^{\top}$, where $\bm{U}$ is $d_1\times d_2$ matrix, $\bm{\Sigma}$ is $d_2\times d_2$ diagonal matrix, $\bm{V}$ is $d_2\times d_2$ matrix. We use $SVD(\bm{A})$ to denote the singular value decomposition of matrix $\bm{A}$ and obtain the corresponding $\bm{U}$, $\bm{V}^{\top}$ and $\bm{\Sigma}$. Let $\sigma_i(\bm{A})$ be the $i$-th largest singular value of $\bm{A}$. For a positive integer $N$, let $[N]=\{1,...,N\}$. For a set $\mathcal{A}$, $|\mathcal{A}|$ denotes the number of elements contained in the set. For theoretical bounds, the notation $\widetilde{O}(\cdot)$ omits logarithmic factors.

\subsection{Stochastic Linear Bandits}
In the MAB model, the learner needs to select an appropriate arm to pull at each round; in stochastic linear bandits (SLB), the expected reward of each arm is a linear function with an unknown parameter $\theta^*\in \mathbb{R}^d$.

For classical linear bandits, the learner is given an arm set $\mathcal{A}=[N]$, which corresponds to known arm vectors $\{\bm{x}^{1},...,\bm{x}^{N}\}\subset\mathbb{R}^{d\times 1}$. In round $t$, the learner chooses an action $\bm{x}_t\in\mathcal{A}$ and receives reward $r_t=\left \langle \bm{x}_t,\bm{\theta}^* \right \rangle+\eta_t$, where $\eta_t$ is conditionally independent of $\bm{x}_t$ given $\bm{x}_{1:t-1},r_{1:t-1}$. For our model, we currently consider SLB in the finite-set setting.

According to \citet{OFUL}, we make the following assumption:
\begin{assumption}
The noise $\eta_t$ is R-sub-Gaussian conditioning on $\bm{x}_{1:t-1},r_{1:t-1}$.
\[
\mathbb{E}[e^{\lambda \eta_t}|\bm{x}_{1:t-1},r_{1:t-1}]\leq exp(R^2\lambda^2/2).
\]
\end{assumption}
\begin{assumption}
$\left \|\bm{\theta}^*  \right \|_2\leq S$, $\left \|\bm{x}  \right \|_2\leq L$ for all action $\bm{x}_t\in \mathcal{A}$, for all $t\in [T]$.
\end{assumption}
The goal of the linear bandits is to collaboratively minimize the cumulative regret defined as
\begin{align}
Regret(T)\nonumber
&:=\sum_{t=1}^{T}\left ( \max_{\bm{x}\in \mathcal{A}}\left \langle\bm{x},\bm{\theta}^*\right \rangle-\left \langle\bm{x}_t,\bm{\theta}^*\right \rangle \right )
\end{align}
where $\bm{x}^*=\argmax_{\bm{x}\in \mathcal{A}}\left \langle\bm{x},\bm{\theta}^*\right \rangle$ is the optimal arm.

We use $\bm{X}_t=[\bm{x}_1,...,\bm{x}_t]^{\top}$ to define the matrix of all actions selected up to round $t$.

\subsection{Computation efficiency}
Although the update overhead of linear bandits can be reduced to $O(d^2)$, such computational cost remains substantial in high-dimensional scenarios, especially in systems emphasizing real-time performance where the per-round computation budget may be strictly constrained. Unlike offline training pipelines that can leverage distributed computing and batching, online bandit algorithms demand immediate responses, and any computation beyond the allocated budget directly translates into service degradation or revenue loss. This practical constraint motivates the development of algorithms that offer a flexible trade-off between computational cost and statistical accuracy, allowing practitioners to adapt to varying system loads and latency requirements.

\subsection{Sketching and the Implicit Low-Dimensional Basis}
\label{sec:basisdim}
In the study of computational overhead in linear bandits, \citet{SOFUL} first proposed Sketched OFUL (SOFUL), which effectively reduces the computational cost of per-round updates via a low-rank sketch $\bm{S}_T\in\mathbb{R}^{l\times d}$ satisfying $\bm{S}_T^{\top}\bm{S}_T\approx\bm{X}_T^{\top}\bm{X}_T$. The goal of sketching is to keep an approximate matrix $\bm{S}_T$ in place of $\bm{X}_T$ for policy updates, so that the $d\times d$ covariance $\bm{V}_t=\lambda\bm{I}+\bm{X}_t^{\top}\bm{X}_t$ is replaced by the cheaper $l\times l$ Gram matrix $\bm{S}_T^{\top}\bm{S}_T$. The classical sketch algorithm Frequent Directions (FD) \citep{FD} truncates the smallest singular value after performing an SVD of the covariance matrix at every update, which inevitably discards information and leads to non-convergence when the spectral tail is heavy. In each round $t$, FD updates as follows:
\begin{align}
\nonumber &[\bm{U},\bm{\Sigma},\bm{V}^{\top}]{=}SVD\left([\bm{S}_{t-1}^{\top};\bm{x}_t]^{\top}\right),\\
\nonumber &\delta_t{=}\sigma_l^2(\bm{\Sigma}), \qquad \bm{S}_{t}{=}\sqrt{\bm{\Sigma}^2-\delta_t\bm{I}}\,\bm{V}^{\top},
\end{align}
where $\sigma_i(\bm{\Sigma})$ is the $i$-th largest singular value of $\bm{\Sigma}$, and $\delta_t$ is the truncated spectral mass. Observe that the sketch update already exposes the structure of a low-dimensional projection: the output $\bm{S}_t$ is of the form $\hat{\bm{\Sigma}}\bm{V}^{\top}$, where $\bm{V}^{\top}$ is an orthonormal basis of a $l$-dimensional subspace and $\hat{\bm{\Sigma}}$ carries the coordinates. In other words, \emph{sketching implicitly maintains a low-dimensional basis of projected coordinates}; its information loss arises not from the basis representation itself, but from the truncation of the spectral mass $\delta_t$ that the projection is forced to discard at every round. This observation motivates a natural question: if we keep the low-dimensional basis but \emph{do not truncate} it, can we preserve the computational benefit of sketching while recovering exact linear regression? This leads us to the orthonormal basis parameterization below.

Rather than approximating the covariance matrix through a lossy sketch, we adopt an \emph{exact} low-dimensional representation based on an orthonormal basis. Consider a set of arms $\mathcal{C}$ whose stacked matrix is $\bm{X}_{\mathcal{C}}\in\mathbb{R}^{|\mathcal{C}|\times d}$, and let $r=\mathrm{rank}(\bm{X}_{\mathcal{C}})$ be its intrinsic rank, a quantity determined solely by the data. We are given a tunable rank bound $l$ that we can set freely. When $l\ge r$, the cluster is exactly low-rank relative to $l$ and the basis can be taken at its full rank $r$ with no information loss; when $l<r$, the rank bound binds and the basis is truncated to $l$ dimensions, in which case the method reduces to the ordinary low-rank (sketch-like) regime. In either case, the \emph{effective dimension} of the basis is
\[
\bar r=\min(r,l),
\]
so that $r$ is controlled by the data while $l$ is controlled by the user. We compute an orthonormal basis $\bm{U}_{\mathcal{C}}\in\mathbb{R}^{d\times \bar r}$ of the $\bar r$-dimensional principal subspace of $\mathrm{range}(\bm{X}_{\mathcal{C}}^\top)$ with $\bm{U}_{\mathcal{C}}^\top\bm{U}_{\mathcal{C}}=\bm{I}_{\bar r}$. Every arm $\bm{x}_i\in\mathcal{C}$ is then represented by its low-dimensional coordinate
\[
\bm{z}_i=\bm{U}_{\mathcal{C}}^\top\bm{x}_i\in\mathbb{R}^{\bar r},
\qquad\text{such that}\qquad \bm{x}_i=\bm{U}_{\mathcal{C}}\bm{z}_i .
\]
When $l\ge r$ (equivalently $\bar r=r$), the basis spans the full column space $\mathrm{range}(\bm{X}_{\mathcal{C}}^\top)$ and the representation incurs \emph{zero} information loss: the regularized Gram matrix in the coordinate space, $\bm{M}_{\mathcal{C}}=\lambda\bm{I}_{\bar r}+\sum_{i\in\mathcal{C}}\bm{z}_i\bm{z}_i^\top$, reproduces the exact cluster covariance through the identity $\bm{U}_{\mathcal{C}}\bm{M}_{\mathcal{C}}\bm{U}_{\mathcal{C}}^\top = \lambda(\bm{I}_d-\bm{U}_{\mathcal{C}}\bm{U}_{\mathcal{C}}^\top)+\bm{X}_{\mathcal{C}}^\top\bm{X}_{\mathcal{C}}$. In this regime the basis parameterization is exact, and is in sharp contrast to FD and its variants, which always truncate the smallest singular value and thereby distort the spectral structure. When $l<r$ (equivalently $\bar r=l$), the basis truncates the residual $r-l$ directions and the parameterization coincides with a standard rank-$l$ sketch, recovering the same spectral error as FD. Throughout the paper we focus on the exact regime $l\ge r$, which is what guarantees regret convergence; we next show how this exact basis parameterization yields a bandit algorithm that is computationally efficient yet provably convergent.

\section{Algorithm}
In this section, we analyze the limitations of existing approaches for improving computational efficiency in linear bandits, and present our proposed algorithm, Coordinate Basis Linear Bandits (CBL). The CBL algorithm not only enhances computational efficiency with a fixed rank bound, but also ensures the convergence of the corresponding regret.
\subsection{Limitations of Existing Approaches}
\textbf{OFUL \citep{OFUL}.} As a classic algorithm for linear bandits, OFUL enjoys rigorous theoretical guarantees on regret convergence. However, its per-round update cost is prohibitively large in high-dimensional settings, which makes it difficult to deploy OFUL in computation-constrained environments.

\textbf{SOFUL \citep{SOFUL} / CBSCFD \citep{CBSCFD}.} SOFUL reduces OFUL's computation to $O(dl)$ via fixed-size matrix sketching. Its regret bound, however, depends critically on the spectral error $\Delta_T$:
\[
\mathrm{Regret}_T^{SOFUL} = \tilde{O}\left((1 + \Delta_T)^{3/2} \left(l + d\log(1 + \Delta_T)\right) \sqrt{T}\right).
\]
When the data exhibit heavy spectral tails, $\Delta_T$ grows with $T$, and the regret collapses to linear. Thus, SOFUL trades statistical reliability for efficiency---a compromise that fails precisely when robust performance is most needed. Although CBSCFD reduces the spectral error term from $O(\Delta_T^{3/2})$ to $O(\sqrt{\Delta_T})$, it still carries the risk of linear regret \citep{DBSLinUCB}.

\textbf{DSBLinUCB \citep{DBSLinUCB}.} DBSLinUCB employs a multi-scale, adaptively adjusted sketch size to ensure that the global spectral error is bounded by a fixed parameter $\varepsilon$, thereby guaranteeing sublinear regret. This effectively addresses the long-standing linear regret issue inherent in SOFUL and CBSCFD. However, DBSLinUCB still has several limitations. First, it maintains gradually growing blocks to adaptively increase the sketch size so as to ensure regret convergence, which makes it inapplicable in computation-constrained scenarios where a fixed rank is required. Second, DBSLinUCB is built on the assumption that the data is low-rank. When the data is full-rank and heavy-tailed, DBSLinUCB degrades to exact rank-1 updates, in which case its time complexity increases to $O(d^2)$---the same as OFUL---thereby losing the speed advantage of low-rank methods.

\subsection{Coordinate Basis Linear Bandits}
We address the regret non-convergence problem in traditional sketching methods and propose the Coordinate Basis Linear Bandits (CBL) algorithm to improve the computational efficiency of updates for linear bandits with a finite arm set, as shown in Algorithm \ref{alg:CBL}. Moreover, our method still guarantees regret convergence even in the presence of full-rank data and under a fixed rank bound. CBL is primarily composed of the following components:

\textbf{Exact Basis Parameterization.} Instead of maintaining a truncated sketch $\bm{S}\in\mathbb{R}^{l\times d}$, each cluster $\mathcal{C}_k$ maintains an orthonormal basis $\bm{U}_k\in\mathbb{R}^{d\times \bar r_k}$ of its column space together with the coordinate matrix $\bm{Z}_k\in\mathbb{R}^{|\mathcal{C}_k|\times \bar r_k}$, where $\bar r_k=\min(r_k,l)$, $r_k=\mathrm{rank}(\bm{X}_k)$ is the intrinsic rank of the cluster, and the row $i$ of $\bm{Z}_k$ is $\bm{z}_i=\bm{U}_k^\top\bm{x}_i$. The regression statistics are then maintained in the coordinate space: the Gram matrix $\bm{M}_k=\lambda\bm{I}_{\bar r_k}+\sum_{s\in\mathcal{C}_k}\bm{z}_{a_s}\bm{z}_{a_s}^\top$ and the linear term $\bm{v}_k=\sum_{s\in\mathcal{C}_k}y_s\bm{z}_{a_s}$. When $l\ge r_k$, the basis is exact and no singular value is truncated, so the estimate $\bm{w}_k=\bm{M}_k^{-1}\bm{v}_k$ and the confidence width $\sqrt{\bm{z}_i^\top\bm{M}_k^{-1}\bm{z}_i}$ coincide pointwise with those of the exact $d$-dimensional ridge regression restricted to the cluster. This is the key difference from FD, which always truncates the smallest singular value and thereby discards information whenever the cluster is not exactly low-rank. When $l<r_k$, the basis truncates the residual directions and coincides with a rank-$l$ sketch; our theoretical guarantees target the exact regime $l\ge r_k$, which the clustering warm-up enforces by construction.

\textbf{Cluster Basis Warm-up.}
We propose a strategy that integrates clustering with the exact basis parameterization, where each cluster maintains its own basis and coordinates to preserve the local covariance structure of arms within the cluster, rather than relying on a global unified approximation. We partition the original arm set into clusters based on rank information, ensuring that the intrinsic rank $r_k$ of the matrix composed of the arm vectors within each cluster does not exceed the user-set bound $l$ (so that the exact regime $l\ge r_k$ holds by construction). During this clustering process, an arm is assigned to a cluster if its inclusion does not cause the cluster's rank to exceed $l$; otherwise, we search for another suitable cluster or form a new cluster consisting solely of that arm. This rank check is performed by tentatively appending the arm to the cluster and testing whether the resulting column-space rank exceeds $l$, which can be done incrementally by the basis-expansion procedure of Algorithm \ref{alg:InsertArm}.

Algorithm \ref{alg:Warm-up} presents our cluster basis partitioning scheme. We use $\mathcal{C}^{all}$ to denote the set of all clusters. Each cluster $\mathcal{C}$ is a disjoint subset of the arm set, and each cluster maintains its own basis $\bm{U}_{\mathcal{C}}$ and coordinates $\bm{Z}_{\mathcal{C}}$. Algorithm \ref{alg:Warm-up} takes an arm set $\mathcal{A}$ as input and returns a partitioned cluster set $\mathcal{C}^{all}$ after $t=N$ rounds.

In the warm-up phase, the basis expansion process itself corresponds to computing the rank within the cluster (i.e. the matrix $\bm{X}_{\mathcal{C},t}$ obtained by concatenating the arms contained in the cluster as row vectors). Since the cluster is constructed so that $l\ge r_{\mathcal{C}}=\mathrm{rank}(\bm{X}_{\mathcal{C},N})$, the basis is exact and spans the full column space, and we have the exact identity $\bm{U}_{\mathcal{C}}\bm{M}_{\mathcal{C}}\bm{U}_{\mathcal{C}}^\top = \lambda(\bm{I}_d-\bm{U}_{\mathcal{C}}\bm{U}_{\mathcal{C}}^\top)+\bm{X}_{\mathcal{C},N}^{\top}\bm{X}_{\mathcal{C},N}$. Consequently, the basis parameterization incurs no information loss inside any cluster, thereby guaranteeing the convergence of regret.

 \begin{algorithm}[!ht]
    \caption{Cluster Basis Warm-up: $CBW(\mathcal{A},l)$}
	\label{alg:Warm-up}
    \textbf{Input:} $\mathcal{A}$, rank bound $l$\\
    \textbf{Initialization:} $\mathcal{C}^{all}=\phi$, Find=False, $t=0$\\
    \textbf{Output:} partition $\mathcal{C}^{all}$, bases $\{\bm{U}_{\mathcal{C}}\}$, coordinates $\{\bm{Z}_{\mathcal{C}}\}$
    \begin{algorithmic}[1]
    \FOR{$\bm{x}\in \mathcal{A}$}
    \FOR{$\mathcal{C}\in \mathcal{C}^{all}$}
    \STATE $\bm{c}\gets\bm{U}_{\mathcal{C}}^\top\bm{x}$;\quad $\bm{\rho}\gets\bm{x}-\bm{U}_{\mathcal{C}}\bm{c}$
    \IF{$\|\bm{\rho}\|\le\varepsilon$ \textbf{or} $\mathrm{rank}(\bm{X}_{\mathcal{C}})+1\le l$}
    \STATE Find=True
    \STATE $\mathcal{C}=\mathcal{C}\cup \{\bm{x}\}$; update $\bm{U}_{\mathcal{C}},\bm{Z}_{\mathcal{C}}$ via Algorithm \ref{alg:InsertArm}
    \STATE break
    \ELSE
    \STATE Find=False
    \ENDIF
    \ENDFOR
    \IF{not Find}
    \STATE $\mathcal{C}_{new}=\{\bm{x}\}$
     \STATE $\bm{U}_{\mathcal{C}_{new}}=\frac{\bm{x}}{\|\bm{x}\|}$;\quad $\bm{Z}_{\mathcal{C}_{new}}=[\|\bm{x}\|]$
     \STATE $\mathcal{C}^{all}.\text{append}(\mathcal{C}_{new})$, $\text{len}(\mathcal{C}^{all}){+}{=}1$
    \ENDIF
    \STATE $t{+}{=}1$
    \ENDFOR
    \RETURN $\mathcal{C}^{all},\{\bm{U}_{\mathcal{C}}\},\{\bm{Z}_{\mathcal{C}}\}$
    \end{algorithmic}
    \end{algorithm}

\begin{algorithm}[!ht]
    \caption{InsertArm$(\bm{x}_{\mathrm{new}}, \mathcal{C}_k)$}
	\label{alg:InsertArm}
    \textbf{Input:} new arm $\bm{x}_{\mathrm{new}}$, target cluster $\mathcal{C}_k$ with basis $\bm{U}_k$, coordinates $\bm{Z}_k$, rank $r_k$
    \begin{algorithmic}[1]
    \STATE $\bm{c}\gets\bm{U}_k^\top\bm{x}_{\mathrm{new}}$;\quad $\bm{\rho}\gets\bm{x}_{\mathrm{new}}-\bm{U}_k\bm{c}$
    \IF{$\|\bm{\rho}\|\le\varepsilon$}
        \STATE append coordinate $\bm{c}$ to $\bm{Z}_k$
    \ELSE
        \STATE $\bm{u}\gets\bm{\rho}/\|\bm{\rho}\|$
        \STATE $\bm{U}_k\gets[\bm{U}_k,\,\bm{u}]$;\quad $r_k\gets r_k+1$
        \STATE append zero column to $\bm{Z}_k$ for existing arms
        \STATE append $[\bm{c};\,\|\bm{\rho}\|]$ to $\bm{Z}_k$
        \STATE $\bm{M}_k\gets\mathrm{blkdiag}(\bm{M}_k,\lambda)$;\quad $\bm{v}_k\gets[\bm{v}_k;0]$
    \ENDIF
    \STATE \textbf{if} $r_k>l$ \textbf{then} re-cluster $\mathcal{C}_k$ via $CBW(\mathcal{A},l)$
    \end{algorithmic}
    \end{algorithm}

In case (a), the arm lies inside the existing column space and the basis is unchanged, so the insertion costs only one projection $O(dr_k)$. In case (b), a single orthonormal direction is appended at a cost of one Gram--Schmidt step $O(dr_k)$; because the new direction $\bm{u}$ is orthogonal to the existing span, all existing coordinates and statistics are left untouched (the appended column is zero), so no re-computation of existing data is required. The final guard re-clusters whenever the cluster's intrinsic rank $r_k$ would exceed the bound $l$: under a static arm set this never triggers (the warm-up already guarantees $l\ge r_k$), but it handles the streaming/dynamic-arm case in which the user-set $l$ binds, i.e. when $l<r_k$ the basis is truncated to the $l$-dimensional principal subspace as discussed in Section~\ref{sec:basisdim}.

\textbf{Confidence Ellipsoid.} Under our cluster basis partitioning strategy, the exact basis parameterization within each cluster incurs no information loss. Consequently, the linear regression estimates $\bm{w}_{k,t}=\bm{M}_{k,t}^{-1}\bm{v}_{k,t}$ within each cluster can accurately identify the optimal arm corresponding to that cluster, and the induced estimate $\hat{\bm{\theta}}_{\mathcal{C},t}=\bm{U}_k\bm{w}_{k,t}$ is pointwise identical to the exact $d$-dimensional ridge estimate. Following \citet{OFUL} and \citet{SOFUL}, we construct the confidence interval as:
\begin{align}
\label{eq:beta}
\beta_t(\delta)=R\sqrt{l\ln(1+\frac{tL^2}{l\lambda})+2\ln\frac{N}{l\delta}}+S\sqrt{\lambda}.
\end{align}
\begin{lemma}
\label{lem:main1}
With probability at least $1-\delta$, for any $0< t\leq T$ and $\mathcal{C}\in \mathcal{C}^{all}$, we have
\[
\left \|\hat{\bm{\theta}}_{\mathcal{C},t}-\bm{\theta}^*  \right \|_{\bm{\bar{V}}_{\mathcal{C},t}}\leq \beta_{t}(\delta),
\]
where $\bm{\bar{V}}_{\mathcal{C},t}=\lambda\bm{I}+\bm{X}_{\mathcal{C},t}^{\top}\bm{X}_{\mathcal{C},t}$ is the exact regularized correlation matrix for cluster $\mathcal{C}$ in round $t$, which is reproduced exactly by $\bm{U}_k\bm{M}_{k,t}\bm{U}_k^\top$ on the cluster's column space.
\end{lemma}

\textbf{Sentinel Strategy.} Combining with the clustering partition, we propose CBL (Algorithm \ref{alg:CBL}) to perform arm selection and update at each round. We assign a sentinel $U(\mathcal{C})$ to each cluster, initialized to positive infinity (or the upper bound on rewards). During subsequent updates, the sentinel $U(\mathcal{C})$ is set to the maximum upper confidence bound of the reward among all arms within that cluster.

At the beginning of each round, the algorithm selects the cluster with the largest sentinel value as the current active cluster $\mathcal{W}_t$, and then performs LinUCB selection within $\mathcal{W}_t$. Specifically, it chooses the current arm $\bm{x}_t$ with the maximum upper confidence bound of the linear estimate to pull. Subsequently, the basis statistics $\bm{M}_k,\bm{v}_k$ are updated to incorporate $\bm{x}_t$ through its coordinate $\bm{z}_t=\bm{U}_k^\top\bm{x}_t$. Note that after warm-up, the intrinsic rank $r_k$ of each cluster is at most the bound $l$ (i.e. $l\ge r_k$), and the selected arm belongs to the current cluster, so the coordinate $\bm{z}_t$ already lies in the existing basis and no basis expansion is needed; the update reduces to a rank-one update of the $\bar r_k\times\bar r_k$ Gram matrix with $\bar r_k=\min(r_k,l)=r_k$, which effectively behaves as a rank-$r_k$ LinUCB update. Owing to the coordinate parameterization, the inverse $\bm{M}_k^{-1}$ can be maintained efficiently via Sherman--Morrison or Cholesky, and the confidence width $\sqrt{\bm{z}_i^\top\bm{M}_k^{-1}\bm{z}_i}$ requires only $O(\bar r_k^2)$ per arm. The per-round computational cost is thus $O(\bar r_k^2)=O(r_k^2)\le O(l^2)$. Compared with the $O(d^2)$ cost incurred by computing the inverse matrix in the traditional OFUL algorithm, CBL achieves a significant reduction in computational overhead when $l<\sqrt{d}$.

The sentinel strategy allows our method to preserve the OFUL policy at the global level, while enabling the adoption of the LinUCB policy within individual clusters after warm-up. The integration of these two strategies constitutes our CBL algorithm. Overall, the CBL algorithm maintains the optimism-in-the-face-of-uncertainty principle globally and executes a rank-$r_k$ LinUCB method inside each cluster, thereby substantially lowering the per-round computational cost without compromising regret convergence.

\begin{algorithm}[!ht]
    \caption{Coordinate Basis Linear Bandits (CBL)}
	\label{alg:CBL}
    \textbf{Input:} $\mathcal{A}$, rank bound $l$, $\delta$\\
    \textbf{Initialization:}  $\mathcal{C}^{all},\{\bm{U}_k\},\{\bm{Z}_k\}\longleftarrow CBW(\mathcal{A},l)$\\
    $\bm{M}_{k,1}=\lambda\bm{I}_{r_k}, \bm{v}_{k,1}=\bm{0}_{r_k}, \bm{w}_{k,1}=\bm{0}_{r_k}, U(\mathcal{C})=+\infty, \forall \mathcal{C}{\in}\mathcal{C}^{all}$\\
    \textbf{Output:}
    \begin{algorithmic}[1]
    \FOR{round $t=1,...,T$}
    \STATE $\mathcal{W}_t=\argmax_{\mathcal{C}\in \mathcal{C}^{all}}U(\mathcal{C})$
    \STATE $\bm{x}_t=\argmax_{\bm{x}\in\mathcal{W}_t}\left\{\bm{z}^{\top}\bm{w}_{k,t}+\beta_t(\delta)\sqrt{\bm{z}^{\top}\bm{M}_{k,t}^{-1}\bm{z}}\right\}$ where $\bm{z}=\bm{U}_k^\top\bm{x}$
    \STATE Play $\bm{x}_t$ and receive $y_t$
    \STATE $\bm{z}_t=\bm{U}_k^\top\bm{x}_t$;\quad $\bm{M}_{k,t+1}=\bm{M}_{k,t}+\bm{z}_t\bm{z}_t^\top$;\quad $\bm{v}_{k,t+1}=\bm{v}_{k,t}+y_t\bm{z}_t$
    \STATE $\bm{w}_{k,t+1}=\bm{M}_{k,t+1}^{-1}\bm{v}_{k,t+1}$
    \STATE $U(\mathcal{W}_t)=\max_{\bm{x}\in\mathcal{W}_t}\left\{\bm{z}^{\top}\bm{w}_{k,t+1}+\beta_{t+1}(\delta)\sqrt{\bm{z}^{\top}\bm{M}_{k,t+1}^{-1}\bm{z}}\right\}$
    \FOR{$\mathcal{C}\in \mathcal{C}^{all}$ and $\mathcal{C}\ne \mathcal{W}_t$}
     \STATE $\bm{M}_{\mathcal{C},t+1}=\bm{M}_{\mathcal{C},t}, \bm{v}_{\mathcal{C},t+1}=\bm{v}_{\mathcal{C},t}, \bm{w}_{\mathcal{C},t+1}=\bm{w}_{\mathcal{C},t}$
    \ENDFOR
    \ENDFOR 
    \end{algorithmic}
    \end{algorithm}

\section{Theoretical Results}
\subsection{Regret Analysis}
\begin{theorem}
\label{the:main}
Given the rank bound $l$, if $\beta_t(\delta)$ is chosen as specified in Equation \ref{eq:beta} and set $\lambda\geq\max\{1,L^2\}$, then with probability at least $1-\delta$, the expected cumulative regret of Algorithm \ref{alg:CBL} is upper bounded by
\begin{align}
    Regret(T)
    &\leq 2\beta_T(\delta)\sqrt{2lN_{\mathcal{A}}^{l}T\ln\left(1+\frac{TL^2}{\lambda l}\right)}\\
    &=\widetilde{O}\left(l\sqrt{N_{\mathcal{A}}^{l}T}\right)
\end{align}
where $N_{\mathcal{A}}^{l}=|\mathcal{C}^{all}|$ is the number of clusters produced by Cluster basis warm-up on arm set $\mathcal{A}$ with rank bound $l$.
\end{theorem}

The regret bound follows from the exactness of the basis parameterization: within each cluster, the estimate $\hat{\bm{\theta}}_{\mathcal{C},t}$ and the confidence width coincide pointwise with the exact $d$-dimensional ridge regression restricted to the cluster, so the standard OFUL analysis applies verbatim after replacing the ambient dimension $d$ by the cluster's effective dimension $\bar r_k=\min(r_k,l)=r_k$ (since the clustering ensures $l\ge r_k$). We establish this equivalence in Section~\ref{sec:equiv}; it is the common basis of both our regret guarantee and our computational saving, as detailed there. Compared with the results of SOFUL/CBSCFD in the finite arm set setting, our regret bound eliminates the influence of the spectral error term $\Delta_T$, rigorously guaranteeing the convergence of regret for any rank bound $l>1$. This leads to an overall improvement in the algorithm's performance.

Since each cluster has a rank of at most $l$, when the arms are processed sequentially according to Algorithm \ref{alg:Warm-up}, each cluster can contain at least $l$ arms. Therefore, we have $N_{\mathcal{A}}^{l}<\left \lceil N/l \right \rceil $. In this case, the regret bound of CBL is obtained as $\widetilde{O}(\sqrt{lNT})$, which implies that in the worst case, the regret of CBL depends on the data scale $N$. Here, $N_{\mathcal{A}}^{l}$ is a quantity determined by the composition of the dataset $\mathcal{A}$ and the rank bound $l$. If the dataset exhibits strong linear correlation, the number of arms contained in each cluster increases, and consequently the number of cluster $N_{\mathcal{A}}^{l}$ becomes substantially smaller. For datasets with a strong linear structure, by choosing an appropriate $l$, it is possible to achieve $l\sqrt{N_{\mathcal{A}}^{l}}<d$, thereby reducing computational overhead while attaining better regret performance than OFUL.

\subsection{Equivalence of the Basis Parameterization}
\label{sec:equiv}
The following identities serve as the common foundation of both our theoretical and our computational analysis: they show that the basis parameterization is \emph{exactly} equivalent to the full-dimensional ridge regression, and at the same time reveal that the algorithm only ever needs the small $r\times r$ quantities $(\bm{z}_i,\bm{M}_k)$ rather than the $d\times d$ matrix $\bm{V}_t^{-1}$. Let $\bm{V}_t=\lambda\bm{I}_d+\sum_s \bm{x}_{a_s}\bm{x}_{a_s}^\top$ be the exact regularized Gram matrix of all played arms, and write $\bm{x}=\bm{U}_k\bm{z}$ together with the pure Gram matrix $\tilde{\bm{M}}_k=\sum_s \bm{z}_{a_s}\bm{z}_{a_s}^\top$ (so that $\bm{M}_k=\lambda\bm{I}_r+\tilde{\bm{M}}_k$ already incorporates $\lambda$). Then
\[
\bm{V}_t=\lambda\bm{I}_d+\bm{U}_k\tilde{\bm{M}}_k\bm{U}_k^\top .
\]
Multiplying on the right by $\bm{U}_k$ and using the column orthonormality $\bm{U}_k^\top\bm{U}_k=\bm{I}_r$,
\[
\bm{V}_t\bm{U}_k=\lambda\bm{U}_k+\bm{U}_k\tilde{\bm{M}}_k(\bm{U}_k^\top\bm{U}_k)
      =\lambda\bm{U}_k+\bm{U}_k\tilde{\bm{M}}_k
      =\bm{U}_k\bigl(\lambda\bm{I}_r+\tilde{\bm{M}}_k\bigr)
      =\bm{U}_k\bm{M}_k .
\]
Left-multiplying by $\bm{V}_t^{-1}$ and right-multiplying by $\bm{M}_k^{-1}$ (valid since $\bm{M}_k\succeq\lambda\bm{I}\succ 0$) yields
\[
\bm{V}_t^{-1}\bm{U}_k=\bm{U}_k\bm{M}_k^{-1} .
\]
Consequently, the two quantities actually used by the algorithm depend only on $(\bm{z}_i,\bm{M}_k)$:
\begin{equation}
\label{eq:score}
\bm{x}_i^\top\hat{\bm{\theta}}_t=\bm{z}_i^\top\bm{M}_k^{-1}\bm{v}_k,
\qquad
\bm{x}_i^\top\bm{V}_t^{-1}\bm{x}_i=\bm{z}_i^\top\bm{M}_k^{-1}\bm{z}_i ,
\end{equation}
where the linear term is expanded as $\bm{\beta}_t=\sum_s y_s\bm{x}_{a_s}=\bm{U}_k\bm{v}_k$. Both quantities are independent of $\bm{U}_k$; therefore $\hat{\bm{\theta}}$ and the confidence width are pointwise identical to the exact solution, the arm choices are identical, and the regret trajectory is unchanged. This is the theoretical side of the equivalence: it justifies applying the OFUL analysis verbatim in the coordinate space.

On the computational side, Equation~\eqref{eq:score} states that the confidence width---the only quantity involving the inverse covariance---can be computed entirely from the $r\times r$ matrix $\bm{M}_k^{-1}$, so CBL never needs to form or invert the $d\times d$ matrix $\bm{V}_t$. This is precisely why the per-round cost drops from $O(d^2)$ to $O(r^2)$, as quantified in the next subsection.

\subsection{Computation Analysis}
The computational cost of CBL mainly consists of two parts: the cost incurred during the warm-up phase (Algorithm \ref{alg:Warm-up}), where the arm set is partitioned into clusters and the per-cluster bases are computed, and the per-round update cost during the subsequent online learning phase (Algorithm \ref{alg:CBL}).

Regarding the per-round update cost in the online learning phase, the analysis in the preceding sections has shown that each update occurs within a cluster whose effective dimension $\bar r_k=\min(r_k,l)=r_k$ (with $l\ge r_k$ enforced by warm-up). The dominant cost arises from the coordinate-space Gram update and inversion, each of which is $O(\bar r_k^2)$ (using Sherman--Morrison) or $O(\bar r_k^3)$ (using a fresh Cholesky), and the intra-cluster arm search, which costs $O(\bar r_k^2)$ per arm. Hence, the per-round update cost is $O(\bar r_k^2)=O(r_k^2)\le O(l^2)$. Since $l$ can be tuned in advance according to the requirements of the environment, the setting $l^2<d$ is readily achievable, and thus CBL achieves a significant improvement in computational efficiency. Note that the coordinate projection $\bm{z}_i=\bm{U}_k^\top\bm{x}_i$ costs $O(d\bar r_k)$ per arm, but because the coordinates are \emph{cached} once after warm-up and reused thereafter, this cost does not appear in the per-round online loop.

Regarding the computational cost of the warm-up phase, it mainly arises from the fact that each arm must be tested against each existing cluster to verify whether it can be added to that cluster. In the worst case, this incurs a total computational cost of $O(N\cdot N_{\mathcal{A}}^{l}\cdot ld)$, where $N_{\mathcal{A}}^{l}$ denotes the number of clusters and each rank test requires a projection costing $O(d\bar r_k)\le O(dl)$. Since both the number of clusters and the rank of each cluster grow gradually during the construction process, the actual computational cost is smaller than this worst-case bound. On the other hand, the additional cost incurred during the warm-up phase is one-time only and does not appear in the subsequent online learning stage. When the horizon $T$ is sufficiently large, this one-time warm-up overhead becomes merely a constant term in the total computational cost. Thus, in terms of total computation, CBL achieves a significant efficiency gain.

\section{Conclusion}
This paper addresses the non-convergence regret caused by cumulative spectral errors in sketching-based linear bandits. We propose Coordinate Basis Linear Bandits (CBL), the first low-rank strategy that achieves convergence under a fixed rank bound, via clustering that partitions the action set, maintains lossless per-cluster orthonormal-basis and coordinate representations, and uses a sentinel to select the optimistic cluster each round. We prove that our method achieves sublinear regret with guaranteed convergence, and reduce per-round computation from $O(d^2)$ to $O(l^2)$.

We acknowledge that our current method still has limitations. Since the warm-up phase of the algorithm relies on reading the entire dataset, CBL is not yet applicable to more general linear contextual bandits with a varying arm set. We plan to extend our approach in future work.

\section{Use of AI Tools}
The authors used AI-assisted tools in the research for this work.

\bibliographystyle{plainnat}
\bibliography{references}

\end{document}